%% file: ICHI_main.tex
\documentclass[10pt,conference,compsocconf,letterpaper]{IEEEtran} 

\usepackage{multicol}
\usepackage{comment}
\usepackage[pdftex]{color, graphicx}
\usepackage{subfig}
\usepackage{amsmath}
\usepackage{amssymb}
\usepackage[vlined]{algorithm2e}
\usepackage{url}
\usepackage{xcolor}
\usepackage{float}
\usepackage[T1]{fontenc}
\usepackage[utf8]{inputenc}
\usepackage{soul}
\usepackage[normalem]{ulem}
\usepackage[space]{cite}

\author{\IEEEauthorblockN{Usha Lokala}
\IEEEauthorblockA{\textit{Dept. of Computer Science} \\
AI Institute, University of South Carolina, USA\\
nlokala@email.sc.edu}

\and
\IEEEauthorblockN{Orchid Chetia Phukan}
\IEEEauthorblockA{\textit{Dept. of CSE} \\
IIIT Delhi, India\\
orchidp@iiitd.ac.in}

\and 
\IEEEauthorblockN{Triyasha Ghosh Dastidar}
\IEEEauthorblockA{\textit{Dept. of Computer Science} \\
BITS Pilani Hyderabad, India\\
f20170829@hyderabad.bits-pilani.ac.in}

\and

\IEEEauthorblockN{Francois Lamy}
\IEEEauthorblockA{\textit{Dept. of Society and Health} \\
Mahidol University, Thailand\\
francois.lam@mahidol.ac.th}

\and

\IEEEauthorblockN{Raminta Daniulaityte}
\IEEEauthorblockA{\textit{College of Health Solutions} \\
Arizona State University, USA\\
raminta.daniulaityte@asu.edu}

\and

\IEEEauthorblockN{Amit Sheth}
\IEEEauthorblockA{\textit{Dept. of Computer Science} \\
AI Institute, University of South Carolina, USA\\
amit@sc.edu}}

\begin{document}

\title{\textit{"Can We Detect Substance Use Disorder?"}: Knowledge and Time Aware Classification on  Social Media from Darkweb}

\maketitle

\begin{abstract}
Opioid and substance misuse is rampant in the United States today, with the phenomenon known as the "opioid crisis". The relationship between substance use and mental health has been extensively studied, with one possible relationship being: substance misuse causes poor mental health. However, the lack of evidence on the relationship has resulted in opioids being largely inaccessible through legal means. This study analyzes the substance use posts on social media with opioids being sold through crypto market listings. We use the Drug Abuse Ontology, state-of-the-art deep learning, and knowledge-aware BERT-based models to generate sentiment and emotion for the social media posts to understand users’ perceptions on social media by investigating questions such as: which synthetic opioids people are optimistic, neutral, or negative about? or what kind of drugs induced fear and sorrow? or what kind of drugs people love or are thankful about? or which drugs people think negatively about? or which opioids cause little to no sentimental reaction. We discuss how we crawled crypto market data and its use in extracting posts for fentanyl, fentanyl analogs, and other novel synthetic opioids. We also perform topic analysis associated with the generated sentiments and emotions to understand which topics correlate with people's responses to various drugs. Additionally, we analyze time-aware neural models built on these features while considering historical sentiment and emotional activity of posts related to a drug. The most effective model performs well (statistically significant) with ($macro  F1$=$82.12$, $recall=$$83.58$) to identify substance use disorder. 


\end{abstract}

\input{ICHI_intro.tex}

\input{ICHI_Background.tex}
\input{ICHI_Data.tex}
\input{ICHI_Methods.tex}
\input{ICHI_Results.tex}
\input{ICHI_Discussion.tex}
\input{ICHI_Ethics.tex}

\bibliographystyle{IEEEtran}
\bibliography{references}
\label{sec:References}

\appendix
\section{Appendix}

\subsection{Social Media Data Source}
\label{sec:source}

We considered six substance use subreddits
\begin{enumerate}
    \item r/drug nerds: \url{https://www.reddit.com/r/drugnerds/},
    \item r/research chemicals: \url{https://www.reddit.com/r/research chemicals/},
    \item r/opiates: \url{https://www.reddit.com/r/opiates/} 
    \item r/heroin: \url{https://www.reddit.com/r/heroin/}
    \item r/suboxone: \url{https://www.reddit.com/r/suboxone/}
    \item r/opiates recovery: \url{https://www.reddit.com/r/piatesRecovery/}
\end{enumerate}





\end{document}

%% file: ICHI_Intro.tex
\begin{figure*}[t!]    
\centering
      \includegraphics[width= 0.75\textwidth, height = 0.28\textwidth]{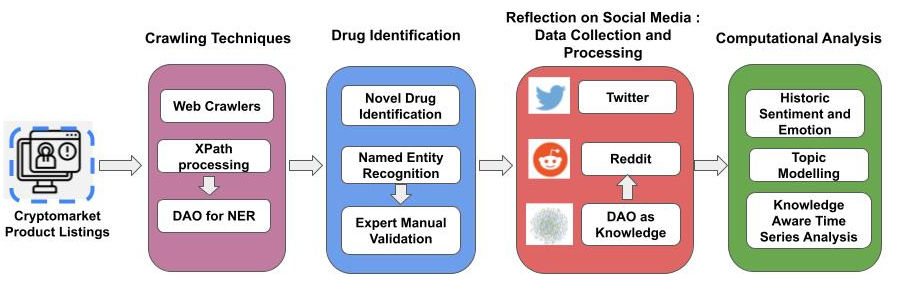}
      \caption{{Proposed Architecture $D2S$ for Harnessing Social Media trends for listings found on Crypto market}}
        \label{fig:fig 1}     
\end{figure*} 
\section{Inroduction}

North America is facing the worst opioid epidemic in its history. This epidemic started with the mass diversion of pharmaceutical opioids (e.g., Oxycodone, Hydromorphone), resulting from the strong marketing advocacy of the potential benefits of opioids \cite{Lamy2020-wx}. The increase in opioid use disorder prevalence and pharmaceutical opioid-related overdose deaths resulted in a stricter distribution of pharmaceutical opioids, unintentionally leading to a dramatic increase in heroin usage among pharmaceutical opioid users \cite{National_Institute_on_Drug_Abuse_undated-wb}. The epidemic entered its third wave when novel synthetic opioids (e.g., fentanyl, U-47,700, carfentanil) emerged on the drug market. Several recent research and reports are pointing at the role of crypto markets in the distribution of emerging Novel Psychoactive Substances (NPS)  \cite{Aldridge2016-oj,National_Academies_of_Sciences_Engineering_and_Medicine2017-wl}. The importance of crypto markets has been further exacerbated by the spillover mental health and anxiety resulting from the ongoing Covid19 pandemic: recent results from the Global Drug Survey suggest that the percentage of participants who have been purchasing drugs through crypto markets has tripled since 2014 reaching 15 percent of the 2020 respondents \cite{GDS2021}.  

In this study, we assess social media data from active opioid users to understand the behaviors associated with opioid usage and to identify what types of feelings are expressed. Substance use disorder (SUD) in social media posts is defined as a post that shows the risk of substance use, attitudes, and behavior related to substance use, as well as the corresponding social and environmental factors \cite{hassanpour2019identifying}. We employ deep learning models to perform sentiment and emotion analysis of social media data with the drug entities derived from crypto markets. We implemented state-of-art sentiment and emotion models for social media data. Also, we performed topic analysis to extract frequently discussed opioid-related topics in social media. For preliminary analysis, we examined temporal variations in topics that  differentiate  between  posts at each drug level and topics over time across all the years, followed by considering data per quarter for each year. We also analyzed how users’ language in their posts varies temporally by topic. We also observed variations in emotions and sentiment that  differentiate  between  posts  containing  expressions of SUD. For this task, we finetuned a pre-trained transformer language model for emotions and sentiments and used it to automatically extract the emotions and sentiments for all the historical posts related to a drug and analyzed variations in sentiment and emotion over time. We further aim to achieve the identification of SUD on social media by examining the core research question of this study: 
\textbf{Can we differentiate between posts containing expressions of substance misuse or not with temporal activity, emotion, sentiment, and language features related to that drug?}
We build a knowledge aware bi-directional sequential neural model that differentiates between posts where expressions of SUD are present versus those posts where it is absent. \\
\textbf{Findings  and  Contributions}
The major contributions and findings of this work are as follows:\\
\begin{enumerate}
\setlength{\itemsep}{0em}
\item
We compile a high-quality, rare, challenging, and valuable dark web dataset (eDark) by crawling four crypto markets namely Dream, Tochka, Agora, and Wall Street. The dataset is available for release upon acceptance.
\item
We propose an end-to-end architecture $D2S$ (Dark web to Social Media) for harnessing social media trends for opioid listings found on the crypto market. It involves Crawling Techniques, Drug identification, Data Collection, Processing from social media, and Computational Models to predict SUD considering the temporal  variations in sentiment and emotional language among posts indicative of SUD. We also contribute the knowledge and historic posts aware sequential neural model that can differentiate if SUD is present or absent for a drug based on these variations by factoring in the relative time difference between historical posts. We  present  that knowledge, sentiment, and emotion-aware models outperform other models of language feature-based approaches by performance measures, ablation study, and error analysis.
\item
To the best of our knowledge, our work is the first one to detect SUD in social media posts considering the above factors and as a reflection of the opioid listings extracted from the dark web. Resources created as a part of the study will be made available upon request to the corresponding author upon acceptance. The resources include emotion, sentiment, and SUD-labeled dataset with timestamps for each drug type, and the $eDark$ dataset.
\end{enumerate}

%% file: ICHI_Background.tex
\section{Related Work}

\textbf{Darkweb Marketplaces:} Darkweb serves as a favorable and promising market for illegitimate goods ranging from drugs to weapons \cite{liggett2020dark, godawatte2019dark, BranwenG}. 
Elbahrawy et al. \cite{elbahrawy2020collective} investigated the market dynamics of dark web markets based on a unique dataset of Bitcoin transactions. They have also analyzed how the market ecology restructures itself once it closes. As traditional web scraping tools have failed to remove the veil of the vendors of dark marketplaces, Hayes et al. \cite{hayes2018framework} proposed an automated framework to overcome this barrier. The suggested framework was further evaluated by gathering information from 3000 sellers on a dark marketplace. Harviainen et al.\cite{harviainen2020drug} presented an analysis of the pattern the buyers and the sellers expose themselves on Sipulitori (Finnish darkweb drug trading market). Hassio et al. \cite{haasio2020information} extended research on Sipulitori by exploring it from the viewpoint of understanding the needs behind the messages posted by users and the physiological and cognitive factors that come into play. Researchers examined the underground marketplaces Agora and Dream Market to examine fluctuations in the availability of fentanyl, fentanyl analogs, and other illegal opioids in connection to overdose fatalities \cite{lokala2019global}. Orsolini et al. \cite{orsolini2017insight} provided intuition behind darkweb drug marketplaces through the perspective of psychiatrists so that they can be equipped with adequate information for providing countermeasures to addiction booming out of drugs available through it. The prior works on analyzing darkweb marketplaces suggest the ability of such data in detecting trends in real world. In the next part of related work, we discuss how time series analysis on social media helps to quantify such trends. 

\par

\noindent\textbf{Time Series Analysis on Social Media:} Earlier research has demonstrated the use of time series analysis on social media data, such as for comprehending changes in public perceptions' sentiment that can be beneficial to the government and commercial organizations \cite{nguyen2012predicting} and understanding the sentiments of users for compelling smartphone applications such as PUBG and TikTok \cite{Iram2020TIMESA}. Time series analysis has also been used in research pertaining to mental health \cite{kolliakou2020mental}, such as variations of mental health of individuals throughout the COVID-19 lockdown phase \cite{pedersen2022time}. \par

Over time, topic analysis and sentiment analysis have been used to deepen the understanding of online retail customer behavior from tweets \cite{ibrahim2019decoding}. Researchers have employed time series analysis to analyze bursts of activity in social networks, and for its prediction, they used LSTM (Long-Short-Term-Memory) network-based model \cite{hajiakhoond2019lstm}. SAGE (Sparse Additive Generative Model), a topic analysis tool, was used to assess the temporal linguistic changes in tweets with and without evidence of self-harm. Furthermore, they explored temporal linguistic features of tweets with and without suicidal intent signs \cite{sawhney2021tweet}. A transformer-based model was also proposed for suicidal ideation detection in social media that takes into consideration the temporal context \cite{sawhney2020time}.
\par
\noindent\textbf{Substance Use Analysis on Social Media:} 
Several researchers have explored social media analysis for different investigations of drug use. These works have analyzed the content, sentiment, and emotion for drug-related data collected from social media platforms like Twitter, and Instagram. Lossio et al. \cite{lossio2018inside} worked on a large amount of opioid-related data collected from Twitter to gain an overall understanding of drug-related discussions on Twitter, behavior related to drug consumption, drugs co-used, and also street terms for various drugs. This study reinstated that Twitter had a huge corpus of data and could provide insights into its correlation with pain management and alcohol consumption. A similar study by Cherian et al. \cite{cherian2018representations}  was conducted on Instagram data on the misuse of codeine. The temporal data collected related to codeine misuse showed its interconnection with alcohol and soda consumption. The influence of social media in propagating this imagery increases the risk of normalizing drug use to extremes. Kim et al. \cite{kim2017scaling} further explored how big data can be utilized to understand drug use and addiction better. Social media is a huge platform for monitoring prescription drug use and addiction using linguistic and behavioral cues. The work done by Lokala et al. \cite{lokala2019global} investigates the relation between the availability of fentanyl-related drugs on crypto markets on the dark web and overdoses of fentanyl. Time-lagged correlation analysis was done between fentanyl-related drugs from the crypto market and overdoses of fentanyl in this first-of-its-kind study for epidemiological surveillance. Sarker et al. \cite{sarker2022concerns} investigated various opioid-related subreddits to better understand the differences in conversations concerning prescription/illegal opioids and access to SUD treatment during the Pre-Covid-19 and Covid-19 periods. They also noticed a rise in opioid withdrawal discussions during Covid-19. Posts from various subreddits related to opioids (both medication and illicit) were collected for identifying the increase in the use of stimulants among opioid users and individuals suffering opioid use disorder \cite{sarker2022trends}. This further corresponds to the increasing number of casualties because of opioids and simulants overdose. Desrosiers et al. \cite{desrosiers2019ru} reported the perseverance of negative sentiments in the conversations of individuals with increased drug use severity. Liu et al. \cite{liu2022linguistic} outlined the presence of positive emotion in Facebook posts of individuals who underwent SUD treatment for a longer period of time than those who stopped their therapy. A study was also made by Singh et al. \cite{singh2021sentiment} to probe the sentiment patterns of tweets related to SUD before and during the Covid-19 pandemic. Cameron et al. \cite{cameron2013predose} followed the development of a semantic web platform called PREDOSE for harvesting data related to prescription drug use from social media platforms. Supporting several types of content analysis, PREDOSE provides easy access to data for drug use research. Fan et al. \cite{fan2017social} illustrates a new framework called AutoDOA to detect drug addiction behavior from Twitter. This will aid in understanding patterns of drug use and addiction. Eshleman et al. \cite{eshleman2017identifying} discussed how social media can be leveraged for drug recovery. Using linguistic patterns and machine learning algorithms, groups of people more likely to participate in the drug recovery process would be an important step in managing the drug addiction epidemic. Our work aims to build an end-to-end system where we see the reflection of the dark web on social media in terms of trends, sentiment, emotion, and substance use context, which is necessary for timely public health intervention.
\begin{figure*}[t!]    
\centering
      \includegraphics[width=0.75\textwidth, height = 0.5\textwidth]{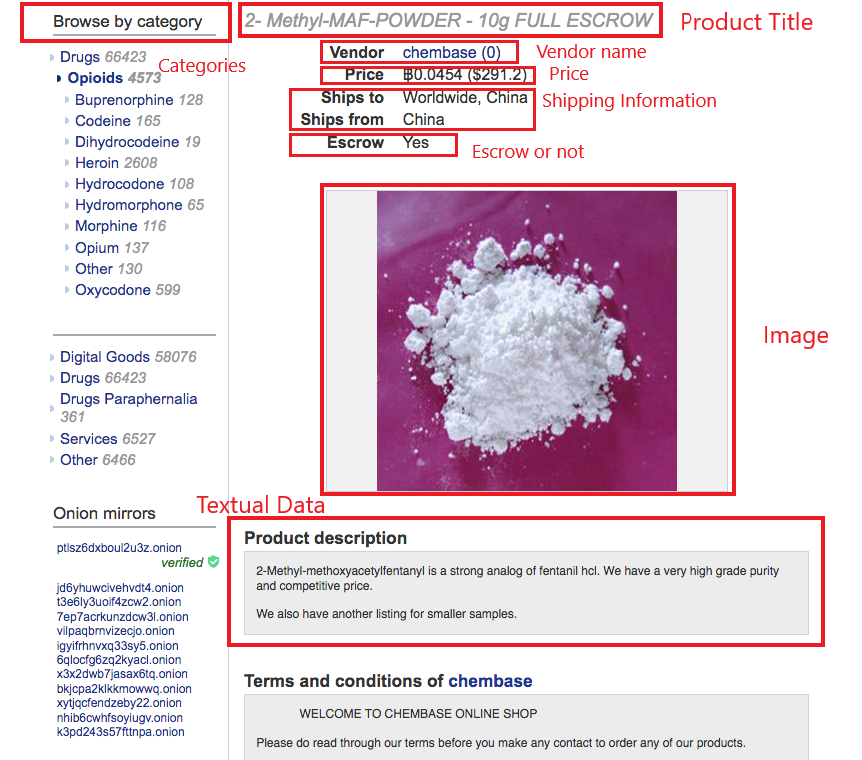}
      \caption{{Data Source of $eDark$: A Sample Product Listing page from Dream Crypto Market}}
        \label{fig:fig 2}     
\end{figure*} 

%% file: ICHI_Data.tex
\section{Data Collection}

This section presents the modules- Crawling Techniques, Drug Identification, and Data collection proposed in $D2S$ architecture as shown in Figure \ref{fig:fig 1}. 
\subsection{eDark Data collection}
Concerning Dark web data, four crypto markets, Agora, Dream market, Tochka, and Wall Street were periodically crawled between June 2014 and January 2020. Over 82,000 opioid-related listings were collected to extract posts about fentanyl, fentanyl analogs and other non-pharmaceutical synthetic opioids in the crypto market. Data sources include four different crypto markets. Further, we discuss $eDark$ dataset summary, and description of crypto markets in this section. 
\subsection{Dark Web Data (eDark) Collection and Summary}
\label{sec:darkweb}
\begin{enumerate}
\setlength{\itemsep}{0em}
\item \textbf{Dream Market}: Late 2013 saw the market's establishment. Dream Market, after AlphaBay, was the largest darknet market in the world prior to 2017. Dream Market quickly overtook AlphaBay as the largest darknet market in the world, nevertheless, once AlphaBay went down in 2017 \cite{thehackernewsDarkUsers}. Between November 2014 and April 2019, there were 261 withdrawals from the market in total. During this time, the market saw transactions worth over 197,000 dollars \cite{whittaker}.
\item \textbf{Tochka}: The market started operating in 2015. It is a fairly modest market that mostly operates in North America and Europe. More than 3621 things, including pharmaceuticals, malware, and other products, are sold on the website. The market changed its name to the Point market and is currently open \cite{smithk}. Between November 2014 and April 2019, there were a total of 2,990 withdrawals from the market. 
\item \textbf{Wall Street}:
The market featured a site for the sale of illegal substances, weapons, hacking tools, and stolen login information. But the exit scam has been hurting the market since April 2019 \cite{darkmarket}. The administrators allegedly stole between 30 million dollar worth of XMR and bitcoins from vendor accounts by switching the site into maintenance mode and transferring the clients' funds \cite{darklaw}. In May 2019, the market was later shut down. Before being taken over in May 2019 by the German Federal Criminal Police, Wall Street was the second-largest darknet market in the entire globe. In total, 7,755 withdrawals were made from the market between November 2014 and April 2019. During this time, there was almost 18,000 dollars worth of transactions on the market \cite{whittaker}.
\item\textbf{Agora}: Agora Market was a darknet market in operation from September 2013 to August 2015 that sold illegal narcotics and controlled
substances, drugs, counterfeit and fraud-related goods, services, and other illegal contraband. The data for Agora for the period June 2014 to September 2015 is obtained from Grams dataset \cite{BranwenG}.

\end{enumerate}

\begin{table}
\centering\begin{tabular}{p{2cm}|p{1cm}|p{1cm}|p{1cm}|p{1cm}} 

 \hline
 \textbf{Data} & \textbf{Dream} & \textbf{Tochka} & \textbf{Wall Street} & \textbf{Agora} \\ 
 \hline
\# Vendors & 3456 &765 & 876 & 910 \\
\hline
\# Substances & 2862 &679 & 765 & 821 \\
\hline
\# Locations & 436 &62 & 37 & 214 \\
\hline
USD Worth & \$ 197k & \$ 5,072 & \$ 18k & \$ 220k \\
\hline
\# Withdrawals & 262 & 2990 & 7755 & 844 \\
\hline
\end{tabular}
\caption{eDark Summary, USD Values and Withdrawals are approximated to nearest value.}
\label{tab: tab 8}

\vspace{-0.5cm}
\end{table}

Agora was chosen because it was one of the largest crypto markets that emerged after the FBI shut down Silk Road \cite{wiredDrugMarket}.
The summary of the dark web dataset is shown in Table \ref{tab: tab 8}.

The sample product page of Dream Market is shown in Figure \ref{fig:fig 2}. The Scrapy framework was used to create the unique web crawler for each market, circumventing security protections built into these markets. To get over security safeguards, it uses specialized Scrapy downloader middleware. By creating a Linux virtual machine on AWS running the Tor daemon and Privoxy, the custom crawler was able to reach the Deep Web. The outputs of the crawler are unaltered HyperText Markup Language (HTML) files used for drug advertising. The University's Information Security Office evaluated and approved the data extraction, storage, and access processes, which all adhered to stringent security standards. The information that was extracted from the data included the following: the product name provided by the vendor, the vendor screen pseudonym, the number of sales made by the vendor, and their level of trust, the drug name(s), drug category, the information the vendor provided about the product, the unit, the quantity in stock, the price (in Bitcoin and US dollars), the price per volume, the country/region of origin, the destination country/region, and the security precautions for transactions. We further used custom-built Named Entity Recognition (NER) to extract substance names, product weight, price of the product, shipment information, availability, and administration route as shown in Table \ref{tab: tab 1}. The NER algorithm consists of three key components: (1) the Natural Language ToolKit (NLTK) is used to curate and process text portions from crawled data; (2) the Drug Abuse Ontology (DAO) that serves as a conceptual framework for interconnecting groups of drug-focused lexicons to produce a list of items to be identified and; (3) Regular Expressions which is a sequence of symbols and characters creating a pattern that can be searched in text or a sentence constructed using the DAO selected entities to extract things of interest. 
\vspace{-0.3cm}

\begin{table}[hbt!]
\resizebox{0.48\textwidth}{!}{%
\begin{tabular}{|c|c|} 
 \hline
 \textbf{Property Name} & \textbf{Crypto market Listing Information} \\ 
 \hline
 Has Product Name & 50 Gr ***** Heroin AAA+ With Spots Free Shipping \\
 \hline
 Is Substance & Heroin \\
 \hline
 Has Class & Opiate \\
 \hline
 Has Dosage & 1.5 gram \\
 \hline
 Has Quantity & 50 gram \\
 \hline
 Has Vendor & BulkBrigade \\
 \hline
 Has Price & BTC 0.0444 \\
 \hline
 Ships To & Worldwide \\
 \hline
 Ships From & Germany \\
 \hline

\end{tabular}}
\caption{Sample of property types in $eDark$ identified from crypto market product listing}
\label{tab: tab 1}
\end{table}

\begin{table*}[]
\begin{tabular}{ |p{4cm}|p{12cm}| } 
 \hline
 \textbf{Opioid Category} & \textbf{Specific Types and subclasses} \\ 
 \hline
Pharmaceutical Fentanyl & Duragesic, Sublimaze, fentanyl transdermal system \\
 \hline
Non-Pharmaceutical Fentanyl & Oxycodone pills with fentanyl \\
 \hline
Fentanyl Analogs & acetylfentanyl, acrylfentanyl, alfentanyl, benzylfentanyl, betahydroxyfentanyl, betamethylfentanyl, butryfentanyl, carfentanil, crotonylfentanyl, etorphine, etorphinecartanil, fluorofentanyl, isobutyrfentanyl, lofentanyl, methoxyacetylfentanyl, methylfentanyl etc.\\
 \hline
 Novel synthetic opioids & U-50488, U-47700, U-49900, U-48800, MT-45, AH-7921, W-18, MPF-47700 etc.\\
 \hline
Pharmaceutical opioids & buprenorphine, codeine, hydrocodone, hydromorphone, loperamide, methadone, morphine, naloxone, oxycodone, oxymorphone, tramadol \\
 \hline
\end{tabular}
\caption{Opioid listings Categories, Subclasses and Specific Types identified using $DAO$}
\label{tab: tab 2}
\end{table*}

\begin{table*}[]
\begin{tabular}{ |p{3cm}|p{12cm}| } 
 \hline
 \textbf{SubReddit} & \textbf{Topics of Interest} \\ 
 \hline
 Opiates Recovery & Cold turkey withdrawal, cravings, anxiety, rehab, depression, sobriety, Loperamide, Benzo, Subutex, quitting, Vivitrol, Imodium, Naltrexone \\
 \hline
 Opiates & Codeine, Hydrocodone, Oxymorphone, Dilaudid, hydromorphone, Opana, Oxycontin, Acetaminophen, Gabapentin, benzos, Roxicodone \\
 \hline
 Suboxone & Buprenorphine, Subutex, Agonist, Clonidine, Tramadol, Hydrocodone, Dilaudid, Vicodin, Sublocade, Percocet, Phenibut, Klonopin, Valium \\
 \hline
 Heroin & Dope, Opium, Opiates, Crack, Diacetylmorphine, China White, codeine, acetaminophen \\
 \hline
 Drug Nerds & Methadone, Alkaloids, Mitragynine, Benzos, Poppy, Buprenorphine, Antagonist, Gabapentin, Naloxone, Amphetamine, Hydrocodone \\
 \hline
 Research Chemicals & Benzos, Psychoactive, Psychedelic, Kratom, Pyrovalerone, Quaalude, Oxycodone, Morphine, Xanax, Tramadol, Cocaine, Methadone, Ketamine, Gabapentin, Amphetamine, Hydromorphone \\
 \hline
\end{tabular}
\caption{Sample of Topics identified from $SUDS$ dataset obtained from six different subreddits }
\label{tab: tab 3}
\end{table*}
\vspace{0.25cm}

\subsection{Named Entity Recognition}
Extracted data included features like product name, vendor screen name (vendor name), drug category, product description, price (Bitcoin or US Dollar), country/region of origin and destination, how to administer the drug, shipping information, and others. We used a pre-trained NER deep learning (NER DL) bidirectional LSTM-CNN approach \cite{Chiu2016-dm} on crypto market data to identify drug entities that use a hybrid bidirectional LSTM and CNN architecture, eliminating the need for most feature engineering. The entities are then matched to a superclass using Drug Abuse Ontology (DAO) \cite{lokala2020dao} that acts as a domain-specific resource with all superclasses related to the entities. DAO is a domain-specific knowledge source containing drug and health-related classes, properties, relationships, and instances. Apart from medical terms, it includes concepts of Mental Health disorders and symptoms aligned with the DSM-5 (Diagnostic and Statistical Manual of Mental Disorders, Fifth Edition) scale. DAO is a domain-specific conceptual framework for interconnecting sets (named “classes”) of drug-focused lexicons. One of the key benefits of using an ontology-enhanced semantic approach is the ability to identify all variants of a concept in data (e.g., generic names, slang terms, scientific names). The DAO contains names of psychoactive substances (e.g., heroin, fentanyl), including synthetic substances (e.g., U-47,700, MT-45), brand and generic names of pharmaceutical drugs (e.g., Duragesic, fentanyl transdermal system) and slang terms (e.g., roxy, fent). It also contains information regarding the route of administration (e.g., oral, IV), unit of dosage (e.g., gr, gram, pint, tablets), physiological effects (e.g., dysphoria, vomiting), and substance form (e.g., powder, liquid, hcl).
Initially, it was used to determine user knowledge, attitudes, and behaviors related to the non-medical use of buprenorphine and other illicit opioids through analysis of web forum data. Later, this ontology evolved to understand trends of drug use in the context of changing legalization policies in the USA. This also proved effective in capturing gleaning trends in the availability of novel synthetic opioids through analysis of crypto market data. DAO is defined utilizing a common ontology methodology known as 101 ontology development. The 101 technique entails the following steps: 1. establishing the ontology's domain and scope; 2. reusing prior knowledge; 3. enumerating key terms in the ontology; 4. defining classes and their properties; and 5. producing instances of the classes. A collection of techniques and best practices accepted by the Semantic Web community and the AI community that do natural language processing were used to assess the ontology's quality. Protege is the most used tool for creating ontologies \cite{Garcia2010-nb}, hence the metrics list the numbers for its structures and representation. The DAO ontology metrics are evaluated as shown in Table \ref{tab: tab 7}.
\\
\vspace{-0.5cm}
\begin{table}[ht]
\centering\begin{tabular}{ p{2.5cm}|p{1cm}|p{4cm}} 

 \hline
 \textbf{Ontology Metric} & \textbf{count} & \textbf{Description} \\ 
 \hline
Axiom & 4876 & No of combined logical and non-logical axioms \\
\hline
Logical Axiom Count & 3478 & No of logical axioms\\
\hline
Declaration Axiom Count & 1185 & No of declaration axioms\\
\hline
 Classes & 316 & No of distinct classes \\
 \hline
Objects & 12 & No of object properties  \\
\hline
Data property & 13 & No of data properties \\
\hline
Individual Count & 845 & No of individual entities \\
\hline
\end{tabular}
\caption{DAO Ontology Metrics}
\label{tab: tab 7}
\end{table}
 
\begin{figure}[t!]    
\centering
 \includegraphics[width=0.3\textwidth, height = 0.5\textwidth]{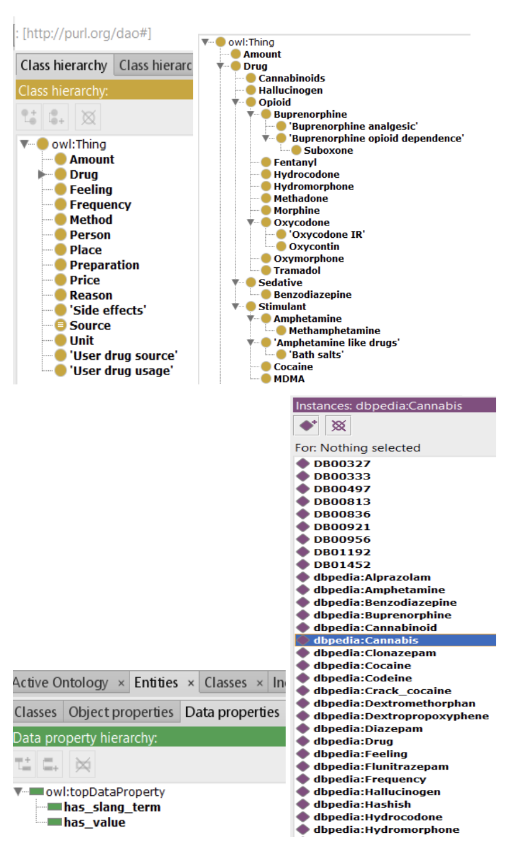}
      \caption{{The OWL (Web
Ontology Language) representation of Drug Abuse Ontology ($DAO$)}}
        \label{fig:fig 3}     
\end{figure}

The OWL (Web Ontology Language) representation of DAO is presented in Figure \ref{fig:fig 3}. In this study, we leverage DAO to identify 90 drug entities, which we then broadly classify into eight categories by mapping each entity to a super drug class in DAO. The eight broad categories considered are Heroin, Synthetic Heroin, Pharmaceutical Fentanyl, Non-Pharmaceutical Fentanyl, Fentanyl, Oxycodone, Kratom, and Opium (chosen as per data available in each category on social media). The categorization of the five types of opioid listings containing specific types and subclasses identified using DAO is shown in Table \ref{tab: tab 2}.
\subsection{Identifying Substance Use Discussions on Social Media}
We crawl the data using a carefully curated lexicon extracted from DAO consisting of around 120 terms (slang names, brand names, drug names, street names, marketing names, commonly used names, abbreviations) of those 8 drug categories. Utilizing the compiled list, we collect 290,458 opioid-related posts from six sub-Reddits using custom-built crawlers, which we call \textbf{S}ubstance \textbf{U}se \textbf{D}isorder corpu\textbf{S} ($\textbf{SUDS}$). The six SubReddits chosen for data collection are r/drug nerds, r/research chemicals, r/opiates, r/heroin, r/suboxone, and r/opiates recovery. The SubReddit sources are mentioned in Appendix \ref{sec:source}.  The SubReddit corpus is spread over different drug categories such as Heroin (136,745), Kratom (77,443), Fentanyl (36,166), Oxycodone (25,890), Opium (9,675), Non-Pharmaceutical Fentanyl (2,798), Pharmaceutical Fentanyl (876), and Synthetic Heroin (865). To build the social media emotion analysis model, additionally, we collected 151,563 posts from Twitter using Twitter API with the same lexicon we used for the subreddit crawl. We applied TF-IDF over unigrams, bigrams, and trigrams to identify topics in each SubReddit as shown in Table \ref{tab: tab 3}. We also conducted the topic analysis using BERTopic \cite{grootendorst2022bertopic} model for all drugs over time from 2015 to 2020, as shown in Figure \ref{fig:overtime}. 

%% file: ICHI_Methods.tex
\section{Methods}

In this section, we build upon the previous data collected to create BERT based Sentiment, Emotion, and SUD models. We leverage those models to predict if a post exhibits substance use disorder present (SUDP) or substance use disorder absent (SUDA) while considering the history of the post. We applied stratified random sampling \cite{hussain2020finite}  to identify a sample population that best represents all the features of interest and ensures that every data subgroup is represented, thus avoiding potential bias in the several datasets we collected for this study.
\subsection{Sentiment Analysis and Sentiment BERT Model}

\begin{figure}[h!]
    \centering
    \subfloat[Pharmaceutical Opioids]
    {\includegraphics[width=0.45\linewidth, height =3.5cm]{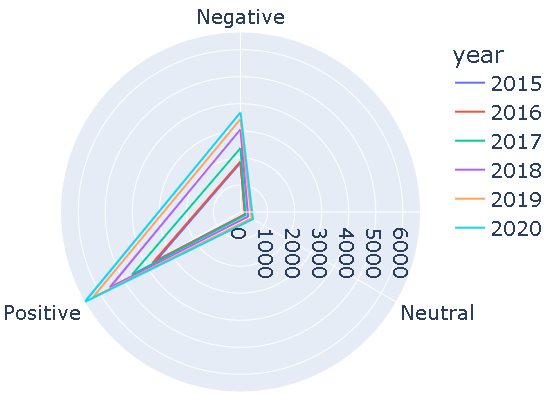}}
    \subfloat[Heroin]
    {\includegraphics[width=0.45\linewidth,height =3.5cm ]{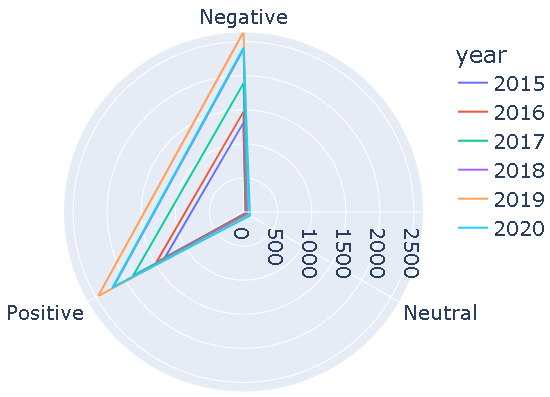}}
   \caption{ $D2S$ - Sentiment Analysis Module - Comparison of drugs Pharmaceutical Opioids and Heroin by sentiments}
\label{fig:sen_time}
\end{figure}

We classified SubReddit posts as Positive, Negative, and Neutral categories for sentiment analysis. We implemented Valence Aware Dictionary for Setiment Reasoning (VADER) \cite{biswas2022drug} to generate sentiment for each SubReddit post in $SUDS$ to consider both the polarity and intensity of each sentiment. VADER uses a lexicon of words with human-annotated sentiment polarity scores like SentiWordNet, AFINN, and the NRC Word-Emotion Association Lexicon. We chose VADER as it is a rule-based sentiment analysis tool that is specifically attuned to sentiments expressed in social media and it uses a combination of sentiment-related words, emoticons, and syntax to produce a sentiment score for a given text. Following the individual scoring of each word, the ultimate sentiment is determined by performing a pooling procedure, such as averaging all the sentiments.  This dataset is split into train, dev, and test sets (75:5:20). The generated training set is used to train state-of-art deep learning algorithms like CNN, LSTM, and BERT. The highest F\textsubscript{1} achieved is 82.36 with the BERT model. We trained the Sentiment BERT model on this training data for later use. We report the statistics of Sentiment labels for SubReddit posts obtained from sampling 800 random data points from each drug category reported in Table \ref{tab: tab 4}. The comparison for drugs Pharmaceutical Opioids and Heroin by
top three sentiments: positive, negative, and neutral for the time period between 2015 and 2020 is presented in Figure \ref{fig:sen_time} which shows the temporal variation in sentiment for each drug.
\label{sec:error}
\vspace{-0.5cm}

\begin{figure*}[hbt!]    
\centering
 \includegraphics[width=0.98\textwidth, height = 0.28\textwidth]{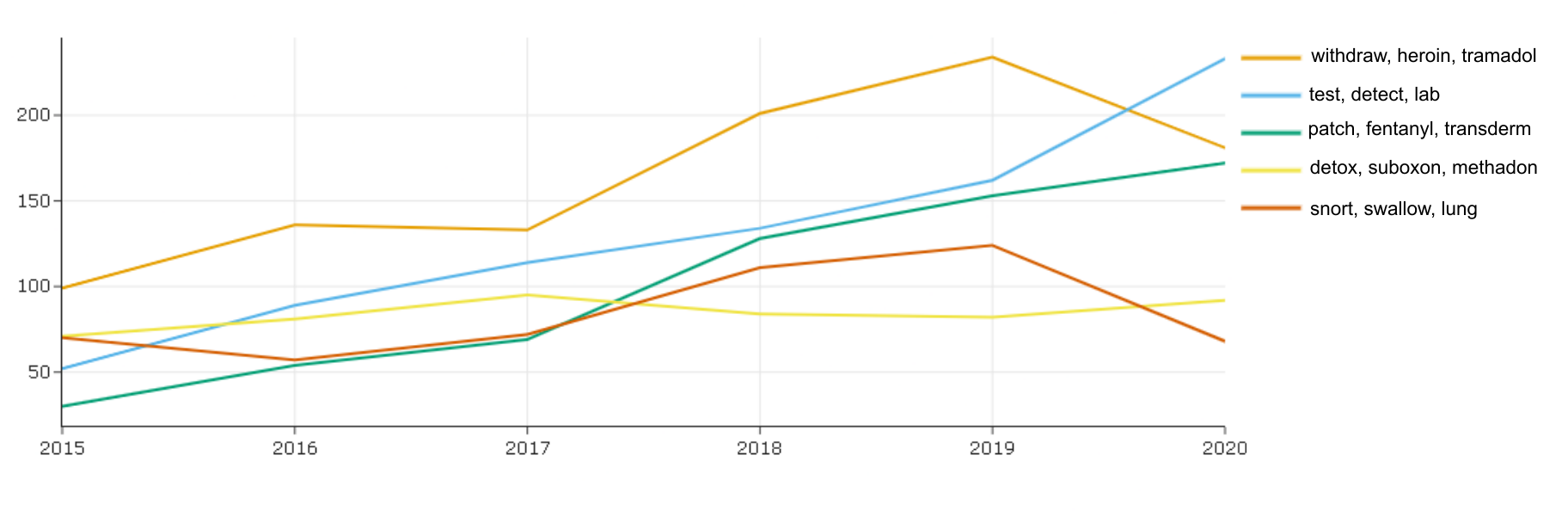}
      \caption{{$D2S$ Topic Modeling Module - Topics over time for all the eight drug categories for time period 2015 - 2020. x axis= Year, y axis= \# of Topics}}
        \label{fig:overtime}     
\end{figure*} 
\begin{table*}[hbt!]
\begin{center}
\begin{tabular}{ |p{2cm}|p{2cm}|p{2cm}|p{2cm}|p{5.5cm}| } 
 \hline
 \textbf{Drug} & \textbf{Positive} & \textbf{Negative} & \textbf{Neutral} & \textbf{Top 3 Emotions in the order found}\\ 
 \hline
Opium & 481 & 218 & 101 & Sadness, Love, Joy \\
 \hline
 Oxycodone & 460 & 245 & 95 & Sadness, Fear, Thankfulness\\
 \hline
 Kratom & 459 & 231 & 110 & Love, Sadness, Fear \\
 \hline
 Fentanyl & 467 & 274 & 59 & Sadness, Love, Fear/Thankfulness \\
 \hline
 Heroin & 455 & 255 & 90 & Sadness, Joy, Thankfulness\\
 \hline
 Synthetic Heroin & 500 & 240 & 6 0 & Sadness, Fear, Thankfulness\\
 \hline
 Pharmaceutical Fentanyl & 570 & 197 & 33 & Sadness, Love, Joy/Thankfulness\\
 \hline
 SNon-Pharmaceutical Fentanyl & 502 & 264 & 34 & Sadness, Love, Thankfulness\\
 \hline

\end{tabular}
\end{center}
\caption{ $D2S$ Sentiment and Emotion Analysis Module - Sentiment stats(Number of Posts) after sampling 800 random points for each drug category identified from 6 Subreddits and Top emotions identified for each drug from Twitter}
\label{tab: tab 4}
\end{table*}
\vspace{0.3cm}

\subsection{Emotion Analysis and Emotion BERT Model}
\begin{figure*}[t!]
    \centering
    \subfloat[Fentanyl]{{\includegraphics[width=0.5\linewidth, height =4.5cm ]{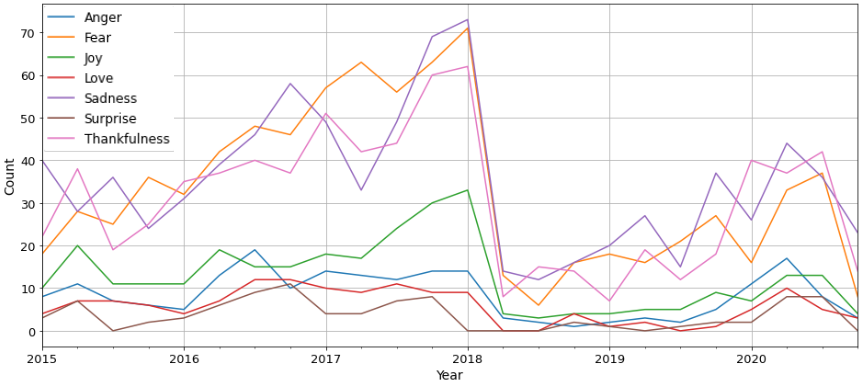}}\label{fig:fen}} 
    \subfloat[Heroin]
    {{\includegraphics[width=0.5\linewidth,height =4.5cm ]{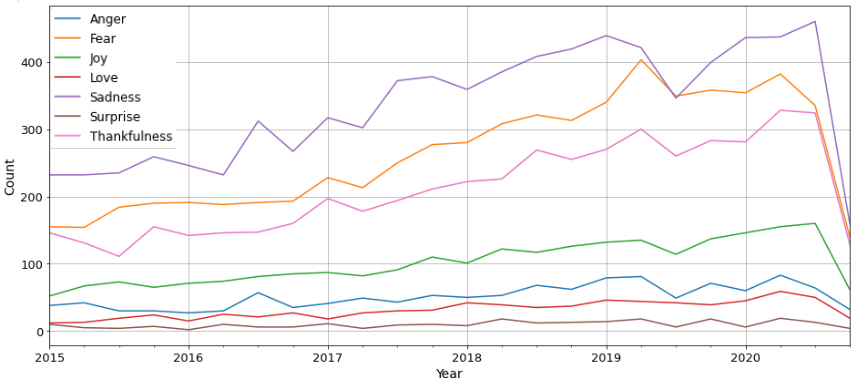}}\label{fig:hero}}\\
    \subfloat[Kratom]{{\includegraphics[width=0.5\linewidth,height =4.5cm ]{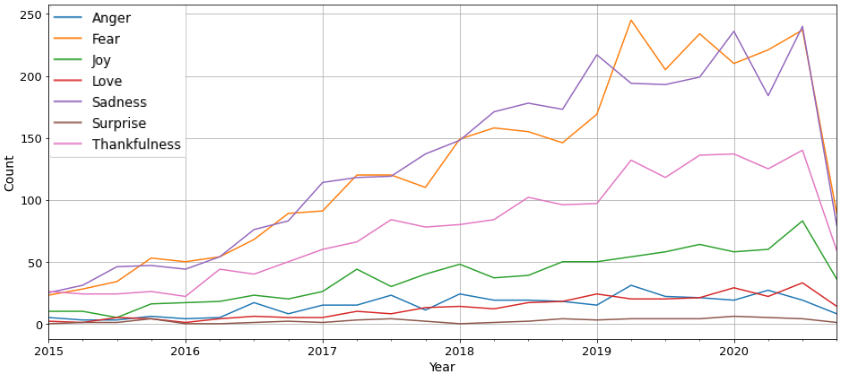}}\label{fig:kra}}
    \subfloat[Pharmaceutical Opioids]{{\includegraphics[width=0.5\linewidth,height =4.5cm]
    {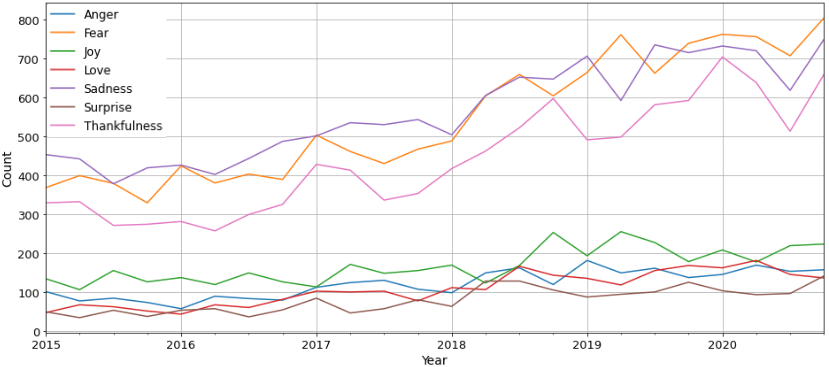}}\label{fig:pop}}\\
    \subfloat[Opium]{{\includegraphics[width=0.5\linewidth,height =4.5cm ]{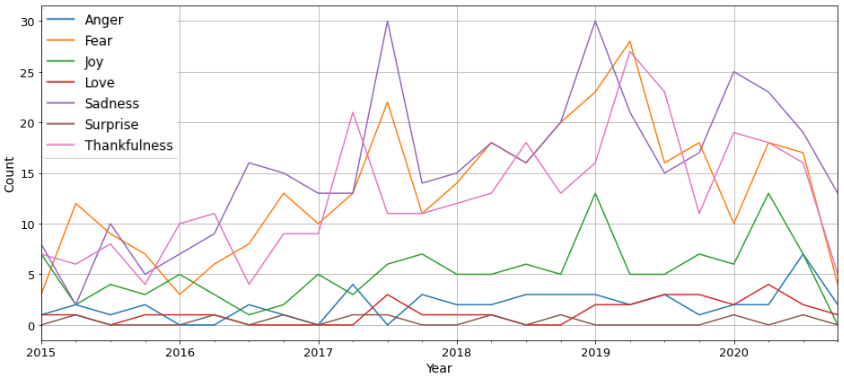}}\label{fig:op}}
    \subfloat[Oxycodone]{{\includegraphics[width=0.5\linewidth,height =4.5cm ]{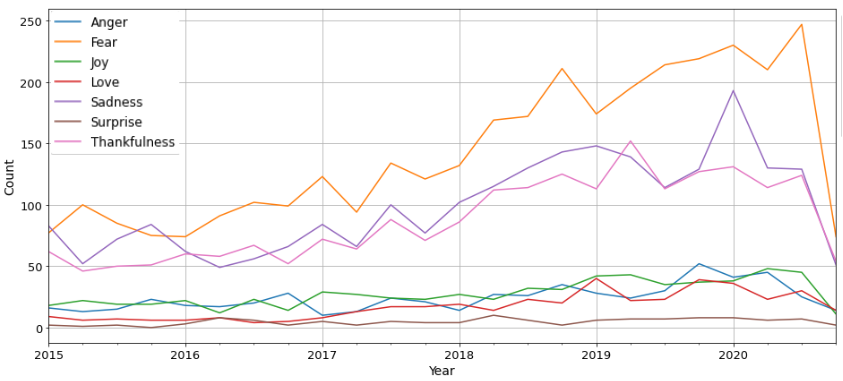}}\label{fig:oxy}}\\
   \caption{ 
    $D2S$ - Emotion Analysis Module - Comparison of eight Major Drug Categories by seven emotions: Joy,  Sadness, Anger, Love, Fear, Thankfulness, and Surprise}
\label{fig:fig 6}
\end{figure*}

We did not choose to work on SubReddit data for emotion analysis as we do not have self-tagged emotions in posts on SubReddit. Therefore, we chose to crawl Twitter for Emotion analysis, where emotions are present as hashtags. We limited our crawl to 7 kinds of emotions, as stated in work done by Wang et al. \cite{Wang2012-bn}. The tweets are assigned a class label corresponding to the emotion hashtag they are associated with. We further remove any URLs or usernames that could potentially contain sensitive information. For generating emotion labels for drug related tweets, we implement an inductive transfer learning approach with BERT \cite{mozafari2019bert}.
For this task, we extracted $61k$ posts as labeled training data by crawling tweets with each emotion hashtag: Joy, Sadness, Anger, Love, Fear, Thankfulness, and Surprise. We split this dataset into train, dev, and test sets (75:5:20). We train Emotion BERT, which is a BERT-based model for 10 epochs using a learning rate of 1e-5,
batch size of 32 on this labeled data and also on Emonet, a corpus of around $790k$ tweets \cite{mohammad2018semeval} to generate the emotion labels for subreddit posts in $SUDS$ corpus.We report the statistics of emotion labels for SubReddit posts obtained from sampling 800 random data points from each drug category reported in Table \ref{tab: tab 4}. The comparison for all drugs by seven emotions: Joy, Sadness, Anger, Love, Fear, Thankfulness, and Surprise is shown in Figure \ref{fig:fig 6}.
\\
\vspace{-0.25cm}

\begin{table}[hbt!]
\centering
\caption{Validation results for Emotion BERT and Substance Use Disorder (SUD) BERT  through the transfer learning approach. The trained model is then used to obtain the emotion labels, SUDP, SUDA labels for the posts in the $SUDS$ Corpus.}
\label{tab: tab5}
\begin{tabular}{llll}
\hline
\textbf{TL-BERT Model} & \textbf{Precision} & \textbf{Recall} & \textbf{F1}\\ \hline
Emotion BERT & $80.12$ & $82.29$ & $81.19$ \\   
SUD BERT & $81.28$ & $83.65$ & $82.44$ \\ \hline
\end{tabular}
\end{table}

\vspace{-0.3cm}

\begin{figure}[t!]    
\centering
      \includegraphics[width=0.5\textwidth, height = 0.4\textwidth]{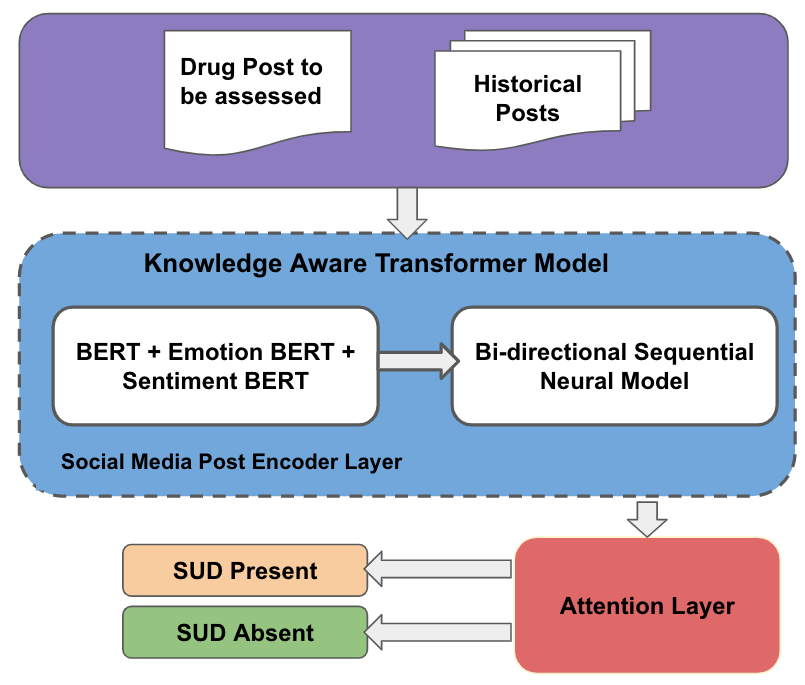}
      \caption{{Knowledge Aware Time Series Analysis Computation Model - A $D2S$ Computational Module}}
        \label{fig:fig 5}     
\end{figure}

\subsection{Substance Use Disorder Dataset}

We focus on building and interpreting a predictive model based on these exploratory results to identify posts where SUD is present or absent. We formulate this problem as a binary classification task to predict a label for a post at a particular time. Each post is associated with a drug name, historical posts, time, emotion, and sentiment. We now prepare our training dataset for generating SUDP and SUDA labels for $SUDS$ corpus. We made use of high-quality addiction-labeled data from Lokala et al. \cite{lokala2021social} work on social media data for exploring the association between drug and mental health symptoms. Lokala et al. \cite{lokala2021social} created a labor-intensive, high-quality corpus of $9888$ tweets manually annotated by domain experts and substance use epidemiologists with experience in Interventions, Treatment, and Addictions Research. We train a Transfer Learning BERT model for 10 epochs using a learning rate of 1e-5 batch size of 32 on this labeled data to generate the SUDP and SUDA labels for posts in the $SUDS$ corpus. We also examine manual inter-annotator agreement among three domain experts for SUDP and SUDA labels of 300 posts which is 0.74 Kappa score, to validate the annotations. The manual annotations are evaluated in the same way as automated labels, and our macro F measure against ground truth is 0.71. The results for transfer learning using BERT are reported in Table \ref{tab: tab5}.

\subsection{Temporal Predictive Model of Posts to detect SUDP or SUDA}
We built our domain-specific Sentiment BERT model to serve as a sentiment feature extractor over historical tweets and a domain-specific Emotion BERT model as an emotion feature extractor for historical tweets. The reason why we built fine-tuned BERT models as they can capture a better sense of sentiment, emotions, social media jargon, and slang terms \cite{wang2022dual}. We contribute Knowledge Aware Time Series Analysis Computation Model to predict SUDP and SUDA for a post as shown in Figure \ref{fig:fig 5}.
We present the SUD detection as a binary classification model with SUDP and SUDA as labels. We focus on building and interpreting a predictive model based on these exploratory results to identify posts where SUD is present or absent. Each post is associated with the drug, historical posts, time, emotion, and sentiment in $SUDS$ corpus.

For a post $P_i$ in SUDS, the concepts/slang terms/synonyms related to drug entities are masked using the DAO ontology, which forms the knowledge component of the model. We use BERT to encode the language representation as BERT can produce more thorough representations of linguistic elements in social media data \cite{wang2022dual}, and we averaged the vector outputs for all tokens in each post at the final layer. To extract emotional language features from posts, We used our Emotion BERT model that takes the historical posts and obtains the 768-dimensional emotion vector of each historical post. To extract sentiment language features from posts, we used our Sentiment BERT model that takes the historical posts and obtains the 768-dimensional sentiment vector of each historical post. Now we have the encoding representing the sentiment and emotion spectrum. Sequential models such as RNN and LSTM models are apt ways to encode representations that learn from a sequence of a user's historical tweets due to the sequential nature of a social media post history. We then pass the historic posts through Bi-LSTM + attention layer concatenated with the post to be assessed. Then we feed the extracted features from the attention layer to a dense layer with the rectified linear unit ($ReLU)$ to get the prediction vector. Finally, we use the softmax function to output the probability that the post has SUDP or SUDA.

For experiments, we split the dataset into 75:5:20 ratios for the train, development, and test sets, respectively. We fine tune the hyper-parameters using the development set. Each model is trained for 10 epochs with a learning rate of 1e-5 batch size of 64. We use cross-entropy loss and ADAM \cite{kingma2014adam} for the optimization. For regularization, we use dropout \cite{hinton2012improving} with a probability of 0.2. We got the best performance with the Bidirectional LSTM (Bi-LSTM) model with the attention layer as it captured context over longer span considering bi-directional context of a word. For all models, we report recall, precision, and F1-score. We interpreted the higher performance gains of our model in the results section through an ablation study.\\

\textbf{Performance Comparisons:} We   compare   the   performance  of  these  state-of-the-art  methods through replications of  the  architectures  and  representations presented in prior works on similar tasks on social media.
\begin{enumerate}
\item
Logistic Regression (LR) \cite{ramkumar2021detecting}: We implement a logistic regression classifier that utilizes part of speech (POS) and term frequency-inverse document frequency (TF-IDF) as language feature representations.
\item
Random Forest (RF) \cite{ji2018supervised}: We implement random forest model with features like Linguistic Inquiry and Word Count (LIWC), POS, and TF-IDF.
\item
History Aware Recurrent neural network (H-RNN) \cite{pitsilis2018effective}: We deploy H-RNN that encodes input using fine-tuned fast text embeddings. Historical posts are passed sequentially through the model and concatenated with the post to be assessed. The sigmoid activation was selected for the the hidden LSTM layer which is fully connected to both the Input and output layer.
\item
History Aware Long Term Short Term Memory (H-LSTM) \cite{cheng2022multimodal}: We replicate H-LSTM that use BERT (Bidirectional Encoder Representations from Transformers) embeddings for encoding historic posts given to an attention based LSTM layer, allowing the model to choose whether to focus more or less on each post in order to reflect user representation finally fed to fully connected layer with a sigmoid activation function to get the prediction. 
\end{enumerate}
\vspace{-0.3cm}

%% file: ICHI_Results.tex
\section{Results}
Out of all Opioid listings in $eDark$, $4.2$\% are novel synthetic opioids, and heroin was identified in $57.8$\% of all opioid-related listings. When comparing the average monthly ad volume for fentanyl, fentanyl analogs, and other non-pharmaceutical drugs, data indicate a rise in the availability of items containing fentanyl. The listings for pharmaceutical and non-pharmaceutical fentanyl and analogs made up $1.9$\% of all opioid-related listings, which is $48.6$\% of unique synthetic opioid-related ads. The most frequent type of novel synthetic opioid which is synthetic heroin is offered for sale at an average of 1.6 kg at each time point of data collection during the study period. Furanylfentanyl was the fentanyl analog that was promoted the most with an average of 3.6 kg being offered for sale at each data point. Carfentanil, a highly strong fentanyl analog was typically available for purchase for 489.6 grams on average. Newer synthetic opioids (e.g., U-48,800, U-4TDP) kept replacing the non-pharmaceutical synthetic opioids (e.g., W-18, MT-45, AH-7921, and U-47,700) in the listings found on marketplaces.

From the exploratory analysis on $SUDS$ corpus, Kratom, Heroin, Fentanyl, Morphine, Cocaine, Methadone, Suboxone, and Oxycodone are the most commonly discussed drugs across six subreddits. In Table \ref{tab: tab 3}, for example, consider Research chemicals (RC); it is interesting to find that more posts talk about Pyrovalerone, a psychoactive drug with stimulant effects. Another term found is ‘Quaalude,’ a brand name for ‘Methaqualone,’ a sedative and hypnotic medication. The RC subreddit mostly discusses psychoactive and psychedelic drugs, while DrugNerds discusses Alkaloids \cite{kaserer2020identification}. Interestingly, DrugNerds talks about Naloxone, which can treat Opioid overdose. Dope is a slang term for Heroin identified in Heroin Subreddit. Several brand names of medications for anxiety, pain, seizures, insomnia, and sedatives are discussed in the Suboxone subreddit. Gabapentin is the typical seizure and pain medication discussed among most of the subreddits. Opiates Recovery is more about withdrawal symptoms and mental health disorders, for example, ‘cold turkey.’ The ‘cold turkey’ used in the context of substance use is quitting a substance abruptly, which carries significant risks if the drug you are discontinuing is benzodiazepine or opiate \cite{Just2016-bk,Landry1992-ge}. The results show that we can derive and analyze slang terms, brand names, novel drugs, mental health symptoms, and medications from social media. 
From the results in Table \ref{tab: tab 4}, it is found that the highest positive sentiment is found in Pharmaceutical Fentanyl, the highest negative sentiment for Fentanyl, and the highest neutral opinion for Kratom. The emotion 'Love' is detected the top one for Kratom as people use it for self-medication. Table \ref{tab: tab 6} presents the medians of metrics for different embeddings and architectures obtained over 20 runs. The baseline models we compared our model with are Logistic Regression (LR), Random Forest (RF), History Aware Recurrent Neural Network (H-RNN), History Aware Long Term Short Term Memory (H-LSTM) with varied language representations like part of speech, term frequency-inverse document frequency. We also extracted LIWC features from posts to pass through a predictive model instead of BERT encoding. Under identical circumstances, we empirically discovered that BERT outperformed LIWC considerably ($p$ < 0.05). We present model interpretability and significance as Ablation Study. 

\begin{table}[hbt!]
\centering
\caption{Ablation Study: Median of metrics over 10 different runs. Bold denotes best performance.}
\label{tab: tab 10}
\begin{tabular}{l|l|l|l}
\hline
\textbf{Model} & \textbf{F\textsubscript{1}-Score} & \textbf{Precision} & \textbf{Recall}  \\ \hline
\textbf{\begin{tabular}[c]{@{}l@{}}Proposed Model\\ (EM+A+H)\end{tabular}} & \textbf{82.12} & \textbf{78.34} & \textbf{83.58} \\ \hline
\textbf{-Attention (A)} & 78.98 (3.14$\downarrow$) & 76.49 (1.85$\downarrow$ ) & 80.34 (3.24$\downarrow$) \\ \hline
\textbf{-Entity Masking (EM)} & 78.50 (3.62$\downarrow$) & 74.61 (3.73$\downarrow$) & 80.15 (3.43$\downarrow$) \\ \hline
\textbf{-History of Post (H)} & 77.58 (4.54$\downarrow$) & 73.72(4.62$\downarrow$) & 79.12 (4.46$\downarrow$) \\ \hline
\end{tabular}
\end{table}

\vspace{1em}

We employ the Wilcoxon Signed Rank Test (Woolson, 2007) to compare the emotional expression in posts and comments between those with and without substance use to assess statistical significance. We see that there is a significant correlation between emotion displayed in posts with SUDP ($p$ < 0.001) and the post with SUDA. We next conduct ablation research, where we remove one component from our model and assess the performance to analyze the prime components in our methodology. Instead of employing attention, we concatenate the substance use post-encoding e\textsubscript{i}\textsuperscript{(S)} and emotion post-encoding e\textsubscript{i}\textsuperscript{(E)} and utilize the resulting representation as input to the linear layer to exclude the attention component from the model. We train our encoders with raw data that is directly collected from social media in order to remove the entity masking component. Also, we trained our model by merely training the classifier and excluding post-history from the model. In Table \ref{tab: tab 10},  we report the findings for the SUD prediction task of posts. Entity masking, which considerably improves the SUD identification task (+3.73 precision, +3.43 recall), is where we see our gains. The Wilcoxon Signed Rank test demonstrates that contextualized representation is very desirable for the SUD identification task in this study since it performs better than the model without entity masking ($p$ < 0.05). Additionally, by adding the history of the post significantly boosts performance where we saw our highest increase (+4.62 precision). Additionally, we see that attention increased the model's precision by $1.85$\% and recall by $3.24$\%, meaning that every feature of the model affects how well it performs this task. Further, We discuss the examples of SUD and the result error Analysis. 

\begin{table*}[hbt!]
\centering
\caption{Examples showing major errors made by our proposed approach.}
\label{tab: tab 9}
\begin{tabular}{p{40em}|c|c|c}    
\hline
\textbf{Post/Comment} & \textbf{Error Type} & \textbf{Actual} & \textbf{Predicted} \\\hline
\textit{1. Imagine a combo. Just got stoned with 12grams of kratom and 15g of **** with 40 mg of ***** AMA lol, ended up projectile vomiting,  went to sleep feeling fearful and woke up feeling pretty joyful but little shitty, so i had glass of ACV, all is well….} & Polydrug use & SUDP & \textcolor{red}{SUDA}\\\hline
\textit{2. The improved memory isn’t dependent on the supplement. I can take the same dose of Nal****one to treat Opioid Use Disorder for 3 months and not feel a thing. That tells me that it helped me heal, and there’s not much more for it to do.} & Post-level Ambiguity & Ambiguous & SUDP \\\hline
\textit{3. why SPEAK the UNSPOKEN? Dangerously F***ing disaster. Mixing *** (a long-term damaging drug, with an ability to make other drugs dangerously stronger) with Xanax ( addictive, overdoseable drug) and ***** (an opiate that can and likely will kill you) is a recipe for life after ***} & Sarcasm Detection & Sarcastic & SUDA \\\hline
\end{tabular}
\end{table*}

\subsection{Error Analysis}

We analyze the sources of errors and discuss the predictions made by our models in Table \ref{tab: tab 9} among three interesting scenarios. 
\begin{enumerate}
\setlength{\itemsep}{0pt}    
    \item \textbf{Polydrug use with variable emotions:} For Post 1, examining the post where multiple drugs co-exist along with emotion variability in history associated with other drugs, for example mixing depressants and stimulants or mixing medications with opioids. Our model is not able to predict correctly. e.g., when substance $A$ might not often co-occur with substance $B$ in history.
    
    \item \textbf{Post-level Ambiguity:} For Post 2, Our model is able to predict SUD by examining the post even if is too ambiguous to assess given the user has clear SUDP in the past, undergoing healing process now with emotion intensity for the historical posts like increased sadness-related emotion.
    
    \item \textbf{Sarcasm detection:} For Post 3, even if it does not contain any clear SUDP/SUDA, but with sarcasm identified in the post, such a post with a history of ambiguous posts presents difficulty identifying SUD which makes it an interesting Natural Language Understanding problem and explains task complexity which levies path for future work.

\end{enumerate}

\begin{table}[!h]
\centering
\begin{tabular}{ |c|c|c|c|c| } 
 \hline
 \textbf{Representation} &\textbf {Model} & \textbf{M-F\textsubscript{1}} & \textbf{P} & \textbf{R} \\ 
 \hline
POS+TF-IDF &LR &52.72 &47.67 &50.31 \\
 \hline
LIWC+POS+TF-IDF &RF & 55.84 &54.63 &56.78 \\
 \hline
Fast Text &H-RNN & 74.47 & 69.48 & 76.17 \\
 \hline
BERT &H-LSTM & 76.85 &70.56 &77.61  \\
 \hline
K-BERT &KH-Bi-LSTM + A & \textbf{82.12} & \textbf{78.34} & \textbf{83.58}\\
 \hline

\end{tabular}
\caption{ Median of Metrics for different embeddings and architectures that were obtained over 20 runs. Bold indicates that the result is significantly better than H-RNN (p<0.05). M-F1 = Macro F1, H = History, LIWC = Linguistic Inquiry and Word Count, POS = part of speech, TF-IDF = term frequency-inverse document frequency, K= Knowledge, A = Attention.}
\label{tab: tab 6}
\end{table}

%% file: ICHI_Discussion.tex
\section{Discussion and Future Work}
Crawling crypto markets poses a significant challenge to apply data science and machine learning to study the opioid epidemic due to the restricted crawling process \cite{Kumar2020-is,Lamy2020-wx,Lokala2020-ay}. To identify the best strategies to reduce opioid misuse, a better understanding of crypto market drug sales that impact consumption and how it reflects social media discussions is needed \cite{Kamdar2019-lm}. We limit this study to eight broad category drugs due to the availability and abundance of related posts on dark web; we hope to refine further and expand our categories for future work. Further, we have identified the processes for future research. We plan to expand this work to extract mental health symptoms from the drug-related social media data to connect the association between drugs and mental health problems, for example, the association between cannabis and depression \cite{Roy2021-yw,Yadav2021-zx}. We also plan to build an Opioid Drug Social Media Knowledge graph with all the diverse data points (Drug, Sentiment, Emotion, mental health symptom, location) and compare it against the state-of-art ‘Knowledge Graph-based Approach For Exploring The U.S. Opioid Epidemic’ \cite{Kamdar2019-lm}. Potential areas of application would be identifying risk factors regarding addiction and mental health from subreddit data \cite{Gaur2018-gj}, and identifying drug trends based on location with a possible Opioid epidemic prediction. We would also like to rely on Drug Enforcement Agency (DEA) Drug Seizures to include in our preliminary data collection process to be aware of related social media discussions.

%% file: ICHI_Ethics.tex
\section{Ethical Statement}
We apply our model to study how historic emotion and sentiment of a drug impacts social media conversation dynamics related to substance use. An important aspect that we need to consider while working with addiction-related issues is to respect the users’ privacy and adhere to good ethical practices adopted by previous research \cite{Henderson2013-nk, Alim2014-bx}. 
Therefore, similar to Matthews et al. \cite{Matthews2017-cn}, we censor several sensitive information, such as user names, personal information, platform identifiers, and URLs which might be directly linked to the user’s identity, from the collected posts. All examples used in this paper are anonymized, and de-identified for user privacy \cite{hunter2018ethical}. We also adopted the proposed guidelines for the treatment of Names and Online Pseudonyms in posts gathered from social media \cite{bruckman2002studying}. 
In this work, we study substance use in Subreddit groups in the form of textual interactions. The expressed addiction intent may differ from the intent actually perceived or experienced by the person. However, obtaining perceived intent from social media is challenging and involves ethical risks. Before behavioral health intervention apps based on social media data can be used in real-world settings, difficulties with potential biases and user privacy must first be resolved, along with establishing suitable regulations and boundaries in this domain. We can adopt the approach used in this work to develop the data set with more human supervision and we acknowledge that the data may be prone to demographic, annotator, and platform-specific biases \cite{preoctiuc2015studying, wang2019demographic}. 
We also acknowledge that the current work does not make any clinical diagnosis or treatment suggestions in any manner whatsoever.